\documentclass[conference]{IEEEtran}
\usepackage{cite}
\usepackage{graphicx}
\graphicspath{ {images/} }
\usepackage{dsfont}
\usepackage{bm}
\usepackage{amssymb}
\usepackage{amsmath}
\usepackage{hyperref}

\title{Predictive Entropy Search for Multi-objective Bayesian Optimization with Constraints}
\author{Eduardo C. Garrido-Merch\'an and Daniel Hern\'andez-Lobato}

\author{
    \IEEEauthorblockN{Eduardo C. Garrido-Merch\'an}
    \IEEEauthorblockA{Universidad Aut\'onoma de Madrid\\
    Francisco Tom\'as y Valiente 11\\
    28049, Madrid\\
    \texttt{eduardo.garrido@uam.es}}
    \and
    \IEEEauthorblockN{Daniel Hern\'andez-Lobato}
    \IEEEauthorblockA{Universidad Aut\'onoma de Madrid\\
    Francisco Tom\'as y Valiente 11\\
    28049, Madrid\\
    \texttt{daniel.hernandez@uam.es}}
}

\date{June 2016}

\begin{document}

\maketitle

\begin{abstract}
This work presents PESMOC, Predictive Entropy Search for  Multi-objective Bayesian Optimization with Constraints, 
an information-based strategy for the simultaneous optimization of multiple expensive-to-evaluate black-box functions 
under the presence of several constraints. PESMOC can hence be used to solve a wide range of optimization problems.
Iteratively, PESMOC chooses an input location on which to evaluate the objective functions and 
the constraints  so as to maximally reduce the entropy of the Pareto set of the corresponding optimization problem. 
The constraints considered in PESMOC are assumed to have similar properties to those of the objective functions in typical Bayesian 
optimization problems. That is, they do not have a known expression (which prevents gradient computation), their evaluation is 
considered to be very expensive, and the resulting observations may be corrupted by noise. These constraints arise 
in a plethora of expensive black-box optimization problems. We carry out synthetic experiments to illustrate 
the effectiveness of PESMOC, where we sample both the objectives and the constraints from a Gaussian process 
prior. The results obtained show that PESMOC is able to provide better recommendations with a smaller number of 
evaluations than a strategy based on random search.
\end{abstract}

\section{Introduction}

We consider the problem of simultaneously minimizing $K$ functions 
$f_1(\mathbf{x}),...,f_K(\mathbf{x})$ subject to the non-negativity 
of $C$ constraint functions $c_1(\textbf{x}),....,c_C(\textbf{x}) $ over some bounded domain 
$\mathcal{X} \in \mathds{R}^d$, where $d$ is the dimensionality of the input space. 
More precisely, the problem considered is:
\begin{align}
\underset{\mathbf{x} \in \mathcal{X}}{\text{min}} & \quad f_1(\mathbf{x}), \ldots, f_K(\mathbf{x}) \nonumber \\
\text{s.t.} & \quad c_1(\mathbf{x}) \geq 0, \ldots, c_C(\mathbf{x}) \geq 0\,.
\end{align}
This scenario is broader and more challenging than the one considered in traditional optimization 
scenarios, where there is a single-objective function and no constraints. In this setting a point 
$\mathbf{x} \in \mathcal{X}$ is feasible if $c_j(\mathbf{x})\geq 0$, for all $j=1,\ldots,C$. 
We define the feasible space $\mathcal{F} \in \mathcal{X}$ as the set of points that are feasible.
Only the solutions contained in $\mathcal{F}$ are considered valid. All potential solutions 
$\mathbf{x}$ not found in $\mathcal{F}$ are ignored.

Most of the times it is impossible to optimize all the objective functions at the same time. 
In particular, they may be conflicting between each other and may prefer different solutions 
$\mathbf{x}\in\mathcal{F}$. An example of this is finding good parameters for the control system 
of a four-legged robot in which we are interested in minimizing energy consumption and maximizing locomotion 
speed \cite{ariizumi2014expensive}. Most probably, maximizing locomotion speed will lead to an increase in the 
energy consumption and vice-versa. In spite of this, it is still possible to find a set of optimal points 
$\mathcal{X}^{\star}$  known as the \textit{Pareto set} \cite{siarry2003multiobjective}. Let us define that $\mathbf{x}$ dominates 
$\mathbf{x}'$ if $f_k (\mathbf{x})\leq f_k (\mathbf{x}')$ $\forall k$, with at least one inequality being strict. 
Then, the Pareto set is the subset of non-dominated points in 
$\mathcal{F}$. Namely, $\forall \mathbf{x}^{\star} \in \mathcal{X}^{\star} \subset \mathcal{F} \,, \forall \mathbf{x} \in \mathcal{F}
\, \exists\, k \in {1,...,K}$ 
such that $f_k(\mathbf{x}^{\star}) < f_k(\mathbf{x})$. Typically, given $\mathcal{X}^{\star}$ the final 
user may choose a point from this set according to their needs (locomotion speed vs. energy consumption). Importantly, 
the Pareto set is often infinite, so most strategies aim at finding a finite set to approximate $\mathcal{X}^{\star}$.

The constraints described in the first paragraph of this section also appear frequently in many optimization
problems. For example, in the problem about the robot described before, without loss of generality, besides optimizing energy 
consumption and locomotion speed, we may have some constraints in the form that the amount of weight placed on 
a leg of the robot does not exceed a specific value, or similarly, that the maximum angle between the legs of 
the robot is below some other value for safety reasons. 

Another example of the problems we are interested in can be found in 
the design process of a new type of low-calorie cookie \cite{gelbart2014bayesian}. In this case the design space may be 
the space of possible recipes and baking times. Here we may be interested in minimizing the number of calories per 
cookie and in maximizing tastiness. These are probably conflictive objectives. Such a problem can also be constrained 
in the sense that we may want to keep production costs below a particular level or we may want that the cookie is 
considered to be crispy for at least 90\% of the population.

The optimization problems considered in this work also arise naturally in the process of tuning machine learning systems. 
Without loss of generality, we may have a deep neural network to be designed for some recognition task and we would like 
to find the architecture and training parameters to simultaneously maximize prediction accuracy and minimize prediction time. These objectives 
are conflictive because reducing prediction error will require bigger and deeper networks which will increase prediction time.
Several constraints may also arise when trying to codify such network in a chip so that it can be included in a low energy 
consuming mobile device. In this case we may want that the energy consumption of the corresponding chip is below a 
particular level. The same may happen with the area of the chip, which could be required to be below some 
particular threshold value.

In many problems of interest the cost of evaluating the objectives or the constraints is very high and
the process may be contaminated by noise. Furthermore, there may be no closed form expressions for 
these function, which will make difficult any gradient computation. This is the case of the examples described 
before. Measuring locomotion speed may involve some experiment with the robot; measuring tastiness may involve some trials with 
some persons; measuring chip energy consumption may involve running a simulator; and so on. If this happens, we would like to account 
for the noise and to minimize the number of evaluations of the objectives and the constraints that is required to obtain the final 
approximation to the Pareto set. An approach that has shown promising results in such a setting consists in using Bayesian 
optimization techniques \cite{mockus1978application}. These techniques use a probabilistic model (typically a Gaussian process 
\cite{rasmussen2006gaussian}) to describe the output of each function. At each iteration, they use the 
uncertainty in the probabilistic models to generate an acquisition function whose maximum indicates the 
most promising location on which to evaluate the objectives and the constraints to solve the optimization 
problem. After enough observations have been collected like this, the probabilistic models can be optimized 
to provide an estimate of the Pareto set of the original problem. Importantly, the acquisition function 
only depends on the uncertainty provided by the probabilistic models and not on the actual objectives or constraints. This 
means that it can be evaluated and optimized very quickly to identify the next evaluation point. 
By carefully choosing the points on which to evaluate the objectives and the constraints, Bayesian optimization methods 
find a good estimate of the solution of the original optimization problem with a small number of evaluations
\cite{brochu2010tutorial,shahriari2016taking}. 

    
In this paper we describe a strategy for constrained multi-objective optimization that is suited to the scenario 
described. For this, we extend previous work that uses information theory to build an acquisition function 
that can be used to optimize several objectives \cite{hernandez2016}, and  also previous work that uses information theory 
to build an acquisition function that can be used to optimize a single objective with several constraints \cite{hernandez2015predictive}. 
The result is a strategy that incorporates the possibility of having several objectives and constraints, 
simultaneously. Such an strategy chooses the next point on which to evaluate the objectives and the constraints as 
the one that is expected to reduce the most the uncertainty about the Pareto set in the feasible space, measured
in terms of \textit{Shannon's differential entropy}. The idea is that a smaller entropy implies that the Pareto set,
\emph{i.e.},  the solution to the optimization problem, is better-identified 
\cite{villemonteix2009informational,hennig2012entropy,hernandez2014predictive}. The proposed approach 
is called Predictive Entropy Search for Multi-objective Bayesian Optimization with Constraints (PESMOC). 

A series of extensive experiments in which both the objectives and the constraints are sampled
from a Gaussian process prior shows that the proposed strategy, PESMOC, has practical advantages over a random 
search strategy that chooses the points on which to evaluate the objectives and the constraints at random.
In particular, PESMOC is able to provide recommendations for the Pareto set that are more accurate with a smaller 
number of evaluations.

\section{Predictive Entropy Search for Multi-objective Optimization with Constraints}

The proposed method maximizes the information gain about the Pareto set $\mathcal{X^{\star}}$ over the feasible set $\mathcal{F}$. 
This method requires a probabilistic model for the unknown objectives and constraints. Let the set of objective 
functions $\{f_1,\ldots,f_K\}$ be denoted with $\mathbf{f}$, and the set of constraint functions 
$\{c_1,\ldots,c_C\}$ be denoted with $\mathbf{c}$. We assume that all these functions have been generated from independent 
Gaussian process (GP) priors \cite{rasmussen2006gaussian}. We also assume observational noise that is i.i.d. Gaussian with zero 
mean. For simplicity, a coupled setting in which all objectives and constraints are evaluated at the same location in 
any given iteration is considered. 

Let $\mathcal{D} = \{(\mathbf{x}_n,\mathbf{y}_n)\}_{n=1}^N$ denote all the observations up to step N, where $\mathbf{y}_n$ is 
a $K+C$-dimensional vector with the values resulting from the evaluation of the $K$ objectives and the $C$ constraints at step 
$n$, and $\mathbf{x}_n$ is a vector in input space representing the corresponding input location. The next evaluation point 
$\mathbf{x}_{N+1}$ is chosen 
as the one that maximizes the expected reduction in the differential entropy $H(\cdot)$ of the posterior distribution over the Pareto 
set in the feasible space, $p(\mathcal{X}^{\star}|\mathcal{D})$. The acquisition function of PESMOC is:
\begin{align}
\alpha(x) & = H(\mathcal{X}^{\star}|\mathcal{D}) - \mathbb{E}_{\textbf{y}} [H(\mathcal{X}^{\star}|\mathcal{D} \cup \{(\textbf{x},\textbf{y})\})]
\label{eq:es}
\,,
\end{align}
where the expectation is taken with respect to the posterior distribution of the noisy evaluations of the 
objectives $\textbf{f}$ and the constraints $\textbf{c}$, at $\mathbf{x}$. That is, 
$p(\textbf{y}|\mathcal{D},\textbf{x}) = \prod_{k=1}^{K}p(y_k|\mathcal{D},\textbf{x}) \prod_{j=1}^C p(y_{K+j}|\mathcal{D},\mathbf{x})$,
under the assumption of independence among the GPs.
In practice, the computation of Eq. (\ref{eq:es}), known as \textit{Entropy Search} \cite{hennig2012entropy}, is very difficult since it 
involves the entropy of a set of points of potentially infinite size. Houlsby \emph{et al.} \cite{houlsby2012collaborative} and 
Hern\'andez-Lobato \emph{et al.} \cite{hernandez2014predictive} describe an approach that makes the computation of that expression easier.
More precisely, Eq. (\ref{eq:es})  represents the mutual information between $\mathcal{X}^{\star}$ and $\textbf{y}$ 
given $\mathcal{D}$. Because the mutual information is symmetric, the roles of $\mathcal{X}^{\star}$ and $\textbf{y}$ can be exchanged 
leading to the following simplified but equivalent expression:
\begin{align}
\alpha(x) & = H(\textbf{y}|\mathcal{D},\textbf{x}) - \mathbb{E}_{\mathcal{X}^{\star}} [H(\textbf{y}|\mathcal{D},\textbf{x},\mathcal{X}^{\star})]
\,,
\label{eq:pes}
\end{align}
where the expectation is now with respect to the posterior distribution for the 
Pareto set in the feasible space, $\mathcal{X}^{\star}$, given the observed data, $\mathcal{D}$,
and $H(\textbf{y}|\mathcal{D},\textbf{x},\mathcal{X}^{\star})$ measures the entropy of 
$p(\mathbf{y}|\mathcal{D}, \textbf{x}, \mathcal{X}^{\star} )$, \emph{i.e.}, the predictive distribution for 
the objectives and the constraints at $\mathbf{x}$ given $\mathcal{D}$ and conditioned to $\mathcal{X}^{\star}$ being 
the Pareto set in the feasible space. This alternative formulation significantly simplifies 
the evaluation of the acquisition function $\alpha(\cdot)$ because we no longer have to evaluate the 
entropy of $\mathcal{X}^{\star}$. We note that the acquisition function obtained in Eq. (\ref{eq:pes}) 
favors the regions of the input space in which $\mathcal{X}^{\star}$ is more informative about 
$\textbf{y}$. These are also the regions in which $\textbf{y}$ is more informative about $\mathcal{X}^{\star}$.

The first term in the r.h.s. of Eq. (\ref{eq:pes}) is straight-forward to evaluate. It is simply the entropy 
of the predictive distribution $p(\mathbf{y}|\mathcal{D}, \textbf{x})$, which is a factorizing $K+C$-dimensional 
Gaussian distribution. That is,
\begin{align}
H(\textbf{y}|\mathcal{D},\textbf{x}) & = \frac{K+C}{2}\log(2\pi e) + \sum_{i=1}^{K}0.5\log(v_{k}^\text{PD}) \nonumber \\
& \quad + \sum_{i=1}^{C}\log(s_c^\text{PD}) \,,
\end{align}
where $v_k^\text{PD}$ and $s_c^\text{PD}$ are the predictive variances of the objectives and the constraints, respectively. 
The difficulty comes from the evaluation of the second term in the r.h.s. of Eq. (\ref{eq:pes}), which is intractable and 
has to be approximated. For this, we follow \cite{hernandez2014predictive,hernandez2016} and approximate the expectation 
using a Monte Carlo estimate obtained by drawing samples of $\mathcal{X}^{\star}$ given $\mathcal{D}$. This involves sampling 
several times the objective functions and the constraints from their posterior distribution $p(\mathbf{f},\mathbf{c}|\mathcal{D})$. 
This is done as in \cite{hernandez2014predictive}. Given a sample of the objectives and the constraints, we 
solve the corresponding optimization problem to find an estimate of $\mathcal{X}^\star$. For this, we use a grid 
search approach, although more efficient methods based on evolutionary strategies can be used in the case 
of high dimensional spaces. The Pareto set $\mathcal{X}^{\star}$ needs to be located in the 
feasible space. Thus, we discard all input grid locations in which the sampled constraints are strictly 
negative. The Pareto set is then simply obtained by returning all the non-dominated 
grid locations.  Note that unlike the true objectives and constraints, the sampled functions can be evaluated very 
cheaply. Given a sample of $\mathcal{X}^{\star}$, the differential entropy of $p(\textbf{y}|\mathcal{D}, \textbf{x}, \mathcal{X}^{\star})$ 
is estimated using expectation propagation \cite{minka2001expectation}.

\subsection{Using Expectation Propagation to Approximate the Conditional Predictive Distribution}

We use expectation propagation (EP) \cite{minka2001expectation} to approximate 
the entropy of the conditional predictive distribution 
$p(\textbf{y}|\mathcal{D}, \textbf{x}, \mathcal{X}^{\star})$. For this, the distribution 
$p(\mathcal{X}^{\star}|\textbf{f},\textbf{c})$ is considered first. 
We note that $\mathcal{X}^{\star}$ is the Pareto set in the feasible space $\mathcal{F}$ if 
and only if $\forall \mathbf{x}^{\star} \in \mathcal{X}^{\star}$, $\forall \textbf{x}'\in\mathcal{X}$, 
$c_j(\mathbf{x}^\star) \geq 0$ for all $j=1,\ldots,C$, and
if $c_j(\textbf{x}') \geq 0$, for all $j=1,\ldots,C$, then $\exists k$ s.t.
$f_k(\mathbf{x}^{\star}) < f_k(\textbf{x}')$. That is, the points in the Pareto set have to 
be feasible and have to be strictly better than any other feasible point in at least one 
of the objectives. Informally, the conditions just described can 
be translated into the following un-normalized distribution:
\begin{align}
\hspace{-.2cm}
p(\mathcal{X}^{\star}|\textbf{f},\textbf{c}) & \propto 
	\prod_{\textbf{x}^\star\in \mathcal{X}^\star} 
	\left(
	\Bigg[\prod_{j=1}^{C}\Phi_j(\textbf{x}^{\star})\Bigg]
	\left[ 
	\prod_{\textbf{x}'\in \mathcal{X}} 
	\Omega(\textbf{x}',\textbf{x}^{\star})
	\right]	
	\right)
	\,,
	\label{eq:x_star}
\end{align}
where $\Phi_j(\mathbf{x}^\star) = \Theta(c_j(\mathbf{x}^\star))$ with $\Theta(\cdot)$ the Heaviside step function, 
and $\Omega(\textbf{x}',\textbf{x}^{\star})$ is defined as:
\begin{align}
\Omega(\textbf{x}',\textbf{x}^{\star}) & = \left[\prod_{j=1}^{C}\Theta(c_j(\mathbf{x}'))\right] 
	\psi(\textbf{x}',\textbf{x}^{\star}) + \nonumber \\
	& \quad \left[1 - \prod_{j=1}^{C}\Theta(c_j(\textbf{x}')) \right] \cdot 1
\,,
	\label{eq:omega}
\end{align}
where $\psi(\textbf{x}',\textbf{x}^{\star})$ is defined as
\begin{align}
\psi(\textbf{x}',\textbf{x}^{\star}) & = 1 - \prod_{k=1}^K \Theta (f_k(\textbf{x}^{\star})-f_k(\textbf{x}'))
\,.  \label{eq:psi}
\end{align}
The factor $\prod_{j=1}^{C}\Phi_j(\textbf{x}^{\star})$ in Eq. (\ref{eq:x_star}) guarantees that every point in the Pareto set 
belongs to the feasible space $\mathcal{F}$ (otherwise $p(\mathcal{X}^{\star}|\textbf{f},\textbf{c})$ is equal to zero). 
The factors $\Omega(\textbf{x}',\textbf{x}^{\star})$ in Eq. (\ref{eq:omega}) 
are explained as follows. The product $\prod_{j=1}^{C}\Theta(c_j(\textbf{x}'))$ checks that the input location $\mathbf{x}'$ 
belongs to the feasible space.  If the point is not feasible, we do nothing, \emph{i.e.}, we multiply everything by one.
Otherwise, the input location $\mathbf{x}'$ has to be dominated by the Pareto point $\mathbf{x}^\star$. That is,
$\mathbf{x}^\star$ has to be better than $\mathbf{x}'$ in at least one objective. That is precisely checked 
by Eq. (\ref{eq:psi}).

Now we show how to compute the conditional predictive distribution $p(\textbf{y}|\mathcal{D}, \textbf{x}, \mathcal{X}^{\star})$.
For simplicity, we consider a noiseless case in which we directly observe the actual objectives and constraints values. 
In that case, $p(\textbf{y}|\textbf{x},\textbf{f},\textbf{c}) = \prod_{k=1}^{K}\delta(y_k - f_k(\textbf{x})) 
\prod_{j=1}^{C}\delta (y_{K+j} - c_j(\textbf{x}))$, where $\delta(\cdot)$ is a Dirac's delta function.  
In the noisy case one only has to replace the delta functions with Gaussians with the corresponding noise variance.
The unnormalized version of $p(\textbf{y}|\mathcal{D}, \textbf{x}, \mathcal{X}^{\star})$ is:
\begin{align}
p(\textbf{y}|\mathcal{D}, \textbf{x}, \mathcal{X}^{\star}) & \propto 
	\int p(\textbf{y}|\textbf{x},\textbf{f},\textbf{c}) 
	p(\mathcal{X}^{\star}|\textbf{f},\textbf{c}) p(\textbf{f}|\mathcal{D}) p(\textbf{c}|\mathcal{D})
 d\textbf{f} d\textbf{c} \nonumber \\
& \propto \int \prod\limits_{k=1}^{K} \delta (y_k - f_k(\textbf{x})) \prod\limits_{j=1}^{C} \delta (y_{K+j} - c_j(\textbf{x}))
	\nonumber  \\
& \quad  \times
	\prod_{\textbf{x}^\star\in \mathcal{X}^\star} 
	\prod_{j=1}^{C}\Phi_j(\textbf{x}^{\star})
	\nonumber \\
& \quad \times
	\prod_{\textbf{x}^\star \in \mathcal{X}^\star}
	\left(
	\Omega(\textbf{x},\textbf{x}^{\star})
	\prod_{\textbf{x}'\in \mathcal{X} \setminus \{\mathbf{x}\}} 
	\Omega(\textbf{x}',\textbf{x}^{\star})
	\right)
	 \nonumber \\
& \quad  \times p(\textbf{f}|\mathcal{D})p(\textbf{c}|\mathcal{D}) d\textbf{f}d\textbf{c}\,,
	\label{eq:exact_acq}
\end{align}
where we have separated the factors $\Omega(\cdot,\cdot)$ that depend on the candidate point $\mathbf{x}$ on which to compute the 
acquisition from the ones that do not depend on $\mathbf{x}$. In order to approximate Eq. (\ref{eq:exact_acq}), using 
EP, $\mathcal{X}$ is approximated with the set 
$\mathcal{X} = \{\textbf{x}_n\}_{n=1}^{N} \cup \mathcal{X}^{\star} \cup \{\textbf{x}\}$.  This set 
represents the union of the input locations where the objectives and constraints have been evaluated, the current Pareto set 
in the feasible space and the candidate point $\mathbf{x}$ where the acquisition function will be evaluated. 
Then, all non-Gaussian factors in Eq. (\ref{eq:exact_acq}) are replaced by Gaussian factors whose parameters are found
using EP \cite{minka2001expectation}. Note that the only non-Gaussian factors are each $\Phi_j(\cdot)$
and each $\Omega(\cdot,\cdot)$.

Non-Gaussian factors are approximated with un-normalized Gaussians as follows.
Each $\Phi_j(\cdot)$ factor is replaced by a one-dimensional un-normalized Gaussian distribution over $c_j(\mathbf{x}^\star)$.
That is,
{\small
\begin{align}
\Phi_j(\mathbf{x}^\star) \approx \tilde{\Phi}_j(\mathbf{x}^\star) \propto
	\exp\left\{-\frac{c_j(\mathbf{x}^\star)^2\tilde{v}_j^{\mathbf{x}^\star}}{2} + 
	c_j(\mathbf{x}^\star) \tilde{m}_j^{\mathbf{x}^\star}\right\}
	\,,
\end{align}
}where $\tilde{v}_j^{\mathbf{x}^\star}$ and $\tilde{m}_j^{\mathbf{x}^\star}$ are natural
parameters adjusted by EP.
Each $\Omega(\mathbf{x}',\mathbf{x}^\star)$ factor is replaced by a product of $C$ one-dimensional
un-normalized Gaussians and $K$ two-dimensional un-normalized Gaussians. In particular,
\begin{align}
\Omega(\mathbf{x}',\mathbf{x}^\star) & \approx
\tilde{\Omega}(\mathbf{x}',\mathbf{x}^\star)  \propto
\prod_{k=1}^K \exp \left\{-\frac{1}{2} \bm{\upsilon}_k^\text{T} \tilde{\mathbf{V}}_k^{\Omega} 
	\bm{\upsilon}_k + (\tilde{\mathbf{m}}_k^{\Omega})^\text{T} \bm{\upsilon}_k \right\} \nonumber \\
	& 
	\times \prod_{j=1}^C 
	\exp\left\{-\frac{c_j(\mathbf{x}^\star)^2\tilde{v}_j^{\Omega}}{2} + 
	c_j(\mathbf{x}^\star) \tilde{m}_j^{\Omega}\right\}
\end{align}
where we have defined $\bm{\upsilon}_k$ as the vector $(f_k(\mathbf{x}'),f_k(\mathbf{x}^\star))^\text{T}$. Furthermore, 
$\tilde{\mathbf{V}}_k^{\Omega}$, $\tilde{\mathbf{m}}_k^{\Omega}$, $\tilde{v}_j^{\Omega}$ and $\tilde{m}_j^{\Omega}$ are
natural parameters adjusted by EP. Note also that $\tilde{\mathbf{V}}_k^{\Omega}$ is a $2 \times 2$
matrix and $\tilde{\mathbf{m}}_k^{\Omega}$ is a two-dimensional vector.

The factors, that do not depend on $\textbf{x}$ (the candidate location on which to compute the acquisition) 
are refined iteratively by EP until they do not change any more. These factors are reused each time that the acquisition 
function has to be computed at a new $\mathbf{x}$. Further details about the EP algorithm are found in \cite{minka2001expectation}. 
The factors that depend on $\mathbf{x}$ are refined only once by EP so that the acquisition function can be 
quickly evaluated. When the EP algorithm finishes, the Conditional Predictive Distribution 
$p(\textbf{y}|\mathcal{D},\textbf{x},\mathcal{X}^{\star})$ of the objectives and constraints at $\mathbf{x}$ is 
approximated by the normalized Gaussian distribution that results from replacing in 
Eq. (\ref{eq:exact_acq}) each non-Gaussian factor by the corresponding Gaussian approximation.
The Gaussian distribution is closed under the product operation, so, when all the factors exposed are replaced 
by Gaussians, the result is a Gaussian distribution.

\subsection{The PESMOC Acquisition Function}

After approximating the Conditional Predictive Distribution by a product of Gaussian distributions using the EP algorithm, 
we add the noise variances to the marginal variances. Then, the PESMOC acquisition function is given by the sum of the differences 
between the entropies before and after conditioning on the Pareto set. This, in combination with the expression shown 
in Eq. (\ref{eq:pes}) gives:
\begin{align}
\alpha(\textbf{x})  & \approx  \sum_{j=1}^{C}\log s_j^{PD}(\textbf{x}) + \sum_{k=1}^{K}\log v_k^{PD}(\textbf{x})
		\nonumber \\
		& \quad - \frac{1}{M}\sum_{m=1}^{M} \Big( \sum_{j=1}^{C}\log s_j^{CPD}(\textbf{x}|\mathcal{X}^{\star}_{(m)}) + 
	\nonumber \\
		& \quad	\sum_{k=1}^{K}\log v_k^{CPD} (\textbf{x}|\mathcal{X}^{\star}_{(m)}) \Big)
	= \sum_{k=1}^K \alpha_k^o(\mathbf{x}) + \sum_{j=1}^C \alpha_j^c(\mathbf{x})\,,
	\label{eq:pesmoc}
\end{align}
where $M$ is the number of Monte Carlo samples, $\{\mathcal{X}^{\star}_{(m)}\}_{m=1}^{M}$, of the Pareto set over the 
feasible set. These samples are used to approximate the expectation in Eq. (\ref{eq:pes}). Furthermore, 
${v}_k^{PD}(\textbf{x})$, ${v}_c^{PD}(\textbf{x})$, ${v}_k^{CPD}(\textbf{x}|\mathcal{X}^{\star}_{(m)})$ 
and ${v}_c^{CPD}(\textbf{x}|\mathcal{X}^{\star}_{(m)})$ are the variances of the predictive 
distribution of the objectives and the constraints, respectively, before and after conditioning on the Pareto set.

We note that the acquisition function in Eq. (\ref{eq:pesmoc}) can be expressed as a sum across the objectives and the constraints. Intuitively,
each term in this sum measures the reduction in the entropy of the Pareto set after an evaluation of the corresponding 
objective or constraint at the input location $\mathbf{x}$. This allows for a decoupled 
evaluation setting in which each objective or constraint is evaluated at a different input location. Furthermore, it 
introduces a mechanism to identify which objective or constraint is expected to be more useful to evaluate.
For this, we only have to individually maximize each of the $K+C$ acquisition functions 
in Eq. (\ref{eq:pesmoc}),
\emph{i.e.}, $\alpha_k^o(\cdot)$, for $k=1,\ldots,K$, and $\alpha_j^c(\cdot)$,  for $j=1,\ldots,C$. 
There is evidence in the literature that a decoupled evaluation setting improves
over a coupled one, for the case of un-constrained multi-objective problems \cite{hernandez2016}.
Similar improvements are expected in the constrained multi-objective case.


The computational cost of evaluating the acquisition function and the EP algorithm is $\mathcal{O}(KCq^3)$, with
$q = N + |\mathcal{X}_{(m)}^{\star}|$, with $N$ being the number of observations, $K$ the number of objectives and $C$ the 
number of constraints. The EP algorithm is run once per each sample of the Pareto set $\mathcal{X}_{(m)}^{\star}$. 
After this, it is possible to re-use again the factors that are independent of the candidate input location $\mathbf{x}$. Thus, 
the complexity of computing the predictive variance is $\mathcal{O}(KC|\mathcal{X}_{(s)}^{\star}|^3)$. In practice, 
the number of points in each Pareto set sample is set equal to $50$, making $q$ to be just a few hundreds at most.

\section{Experiments}

We carry out experiments to compare the performance of the proposed method, PESMOC,
with that of a random search (RS) strategy. At each iteration, in RS the points on which 
to evaluate the objectives and the constraints are obtained by sampling from a 
uniform distribution over the input space. Thus, RS is expected to perform worse than 
PESMOC due to the fact that it does not use the model's uncertainty to identify intelligently
the next point on which to do evaluation. By contrast, PESMOC is expected to use that uncertainty to evaluate
the objectives and the constraints only in those regions of the input space that are expected to be 
more useful. Both strategies, PESMOC and RS, have been implemented in the software 
for Bayesian optimization Spearmint \url{https://github.com/HIPS/Spearmint}.
In our experiments we employ a Mat\'ern covariance function in the GPs that are 
used model the objectives and the constraints. The hyper-parameters of the GPs 
(noise variance, length-scales and amplitudes) are approximately sampled from their posterior distribution 
using slice sampling \cite{NIPS2010_4114}. We generate 10 samples for each hyper-parameter,
and the acquisition function of PESMOC is averaged over these samples.

\begin{figure*}[htb]
\begin{tabular}{cc}
	\includegraphics[width=0.475\linewidth]{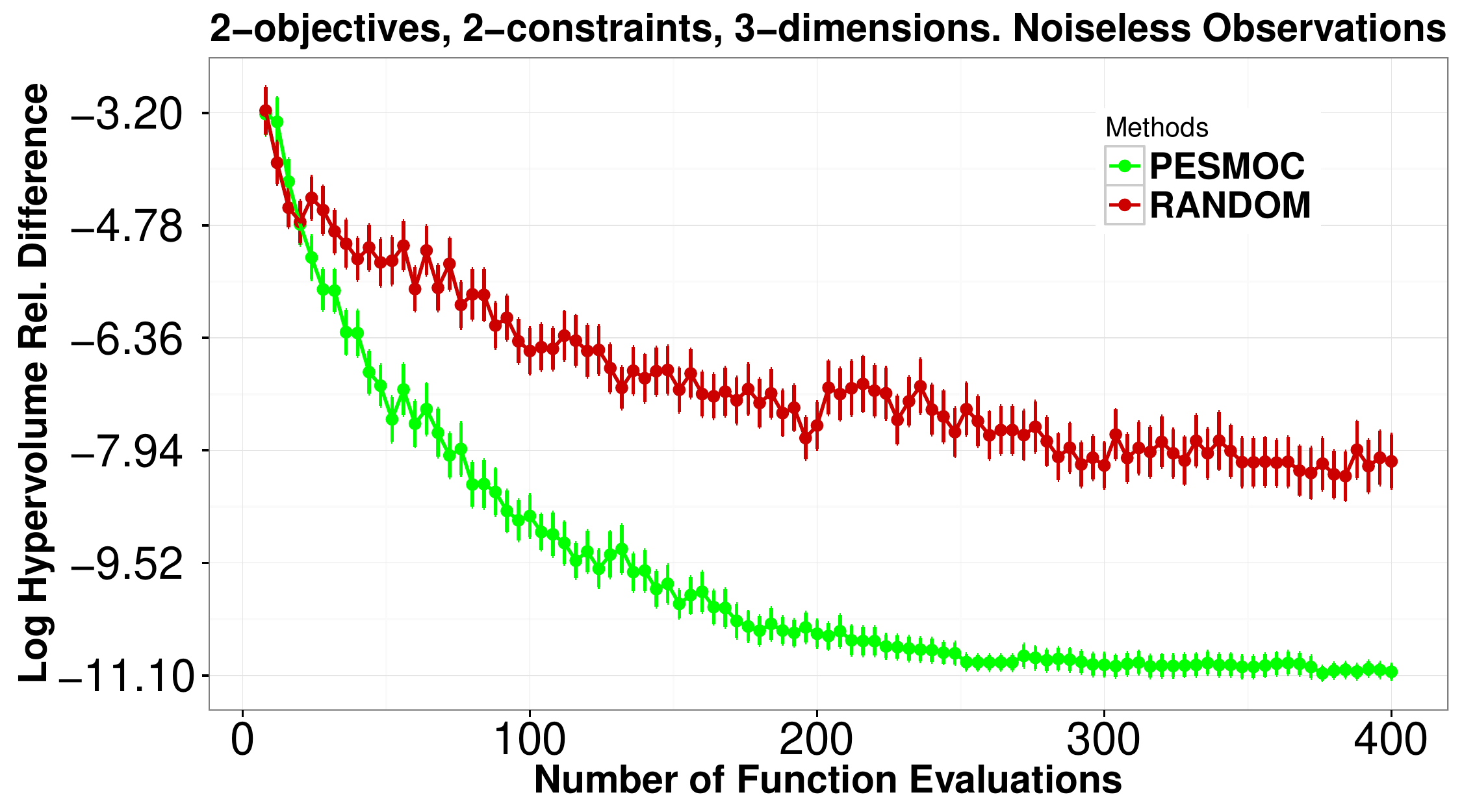} &
	\includegraphics[width=0.475\linewidth]{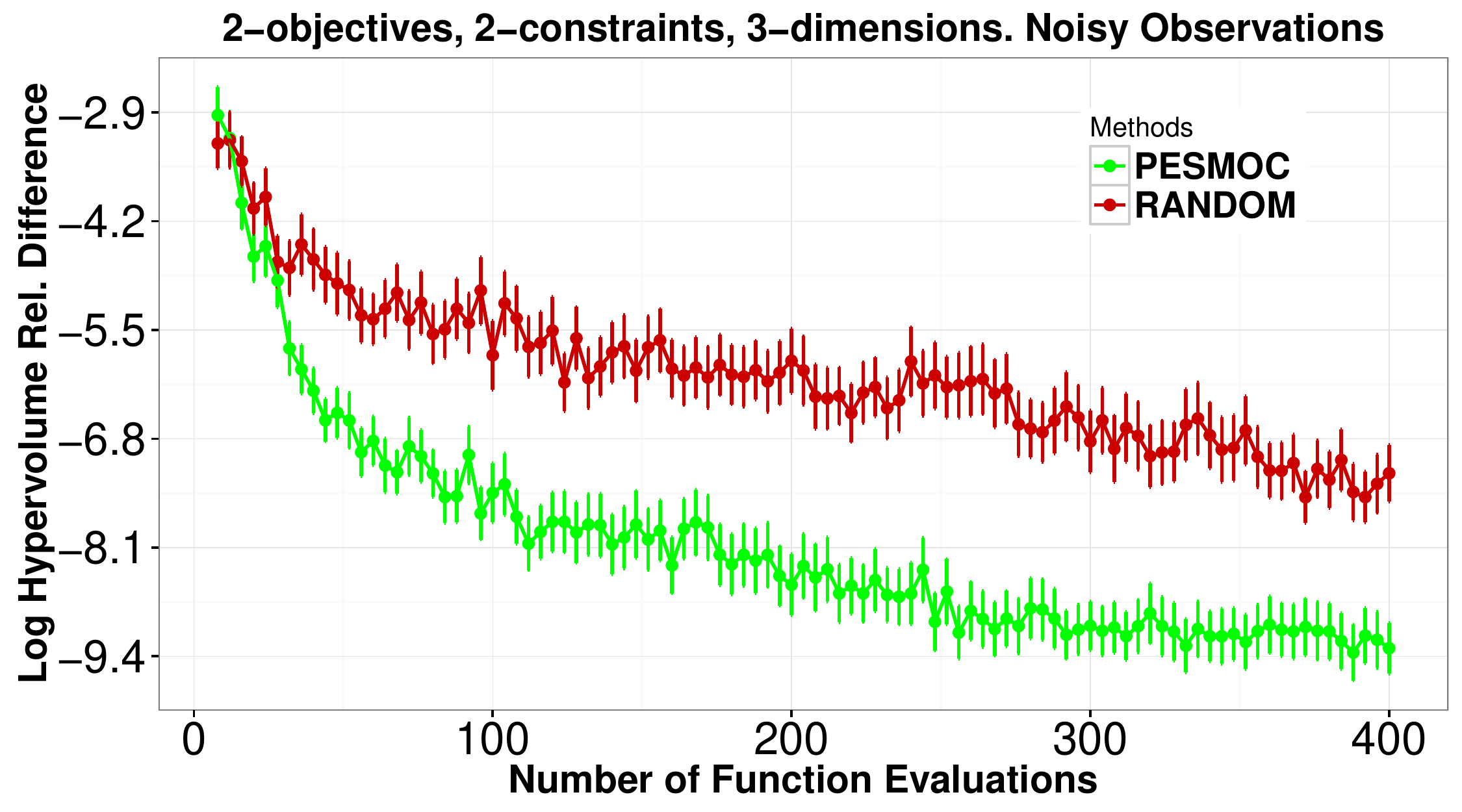}
\end{tabular}
\caption{Logarithm of the relative difference between the hyper-volume of the recommendation obtained by PESMOC
	and the hyper-volume of the recommendation obtained by RS. We report results after reach evaluation of
	the objectives and the constraints. (left) Noiseless evaluations. (right) Evaluations contaminated with
	Gaussian noise. Best seen in color.}
	\label{fig:results_synthetic}
	\vspace{-.5cm}
\end{figure*}

To compare PESMOC and RS, we generate 100 synthetic optimization problems obtained by 
sampling the objectives and the constraints from their respective GP prior. The 
input dimension of each problem is set equal to 3 and we consider 2 objectives and 
2 constraints in each problem. After this, each strategy is run until 100 evaluations
of the objectives and the constraints are done. We consider two scenarios: A first scenario in 
which the evaluations are noiseless, and a second scenario in which 
each evaluation (objectives and constraints) is contaminated with Gaussian noise with 
standard deviation equal to $0.1$. At each iteration, each strategy, PESMOC and 
RS, outputs a recommendation in the form of a Pareto set obtained by optimizing the 
posterior means of the GPs. The performance criterion 
used to compare PESMOC with RS is the hyper-volume indicator, which is maximized by the actual 
Pareto set \cite{zitzler1999multiobjective}. When the recommendation produced by a particular
strategy contains a point which does not belong to the feasible space, we set the corresponding 
hyper-volume equal to zero. In practice, we report the logarithm of the relative difference between 
the hyper-volume of the actual Pareto set and the hyper-volume of the recommendation.

Fig. \ref{fig:results_synthetic} shows the average results obtained in the experiments described, 
alongside with the corresponding error bars. We observe that PESMOC is able to find better solutions 
to the optimization problems considered. More precisely, the solutions obtained by PESMOC are more accurate 
than those obtained by RS, since they have a hyper-volume that is closer to the hyper-volume of the actual Pareto set.
Furthermore, they are obtained with a smaller number of evaluations of the objectives and the constraints. If 
the computational cost of evaluating the objectives and the constraints if very high, this represents a 
significant improvement.

\begin{figure}[htb]
	\begin{tabular}{c}
	\includegraphics[width=0.95\linewidth]{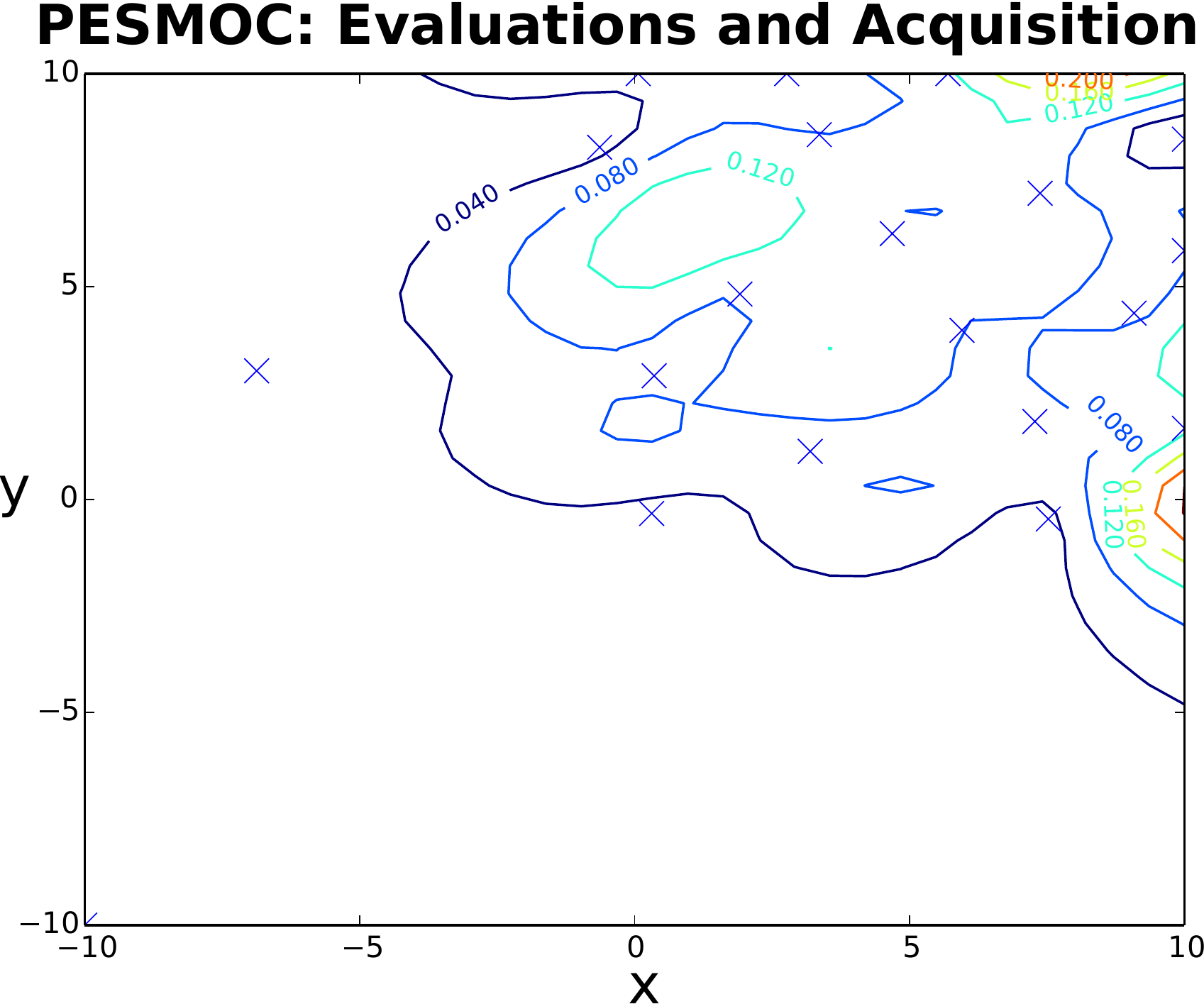}\\
	\includegraphics[width=0.95\linewidth]{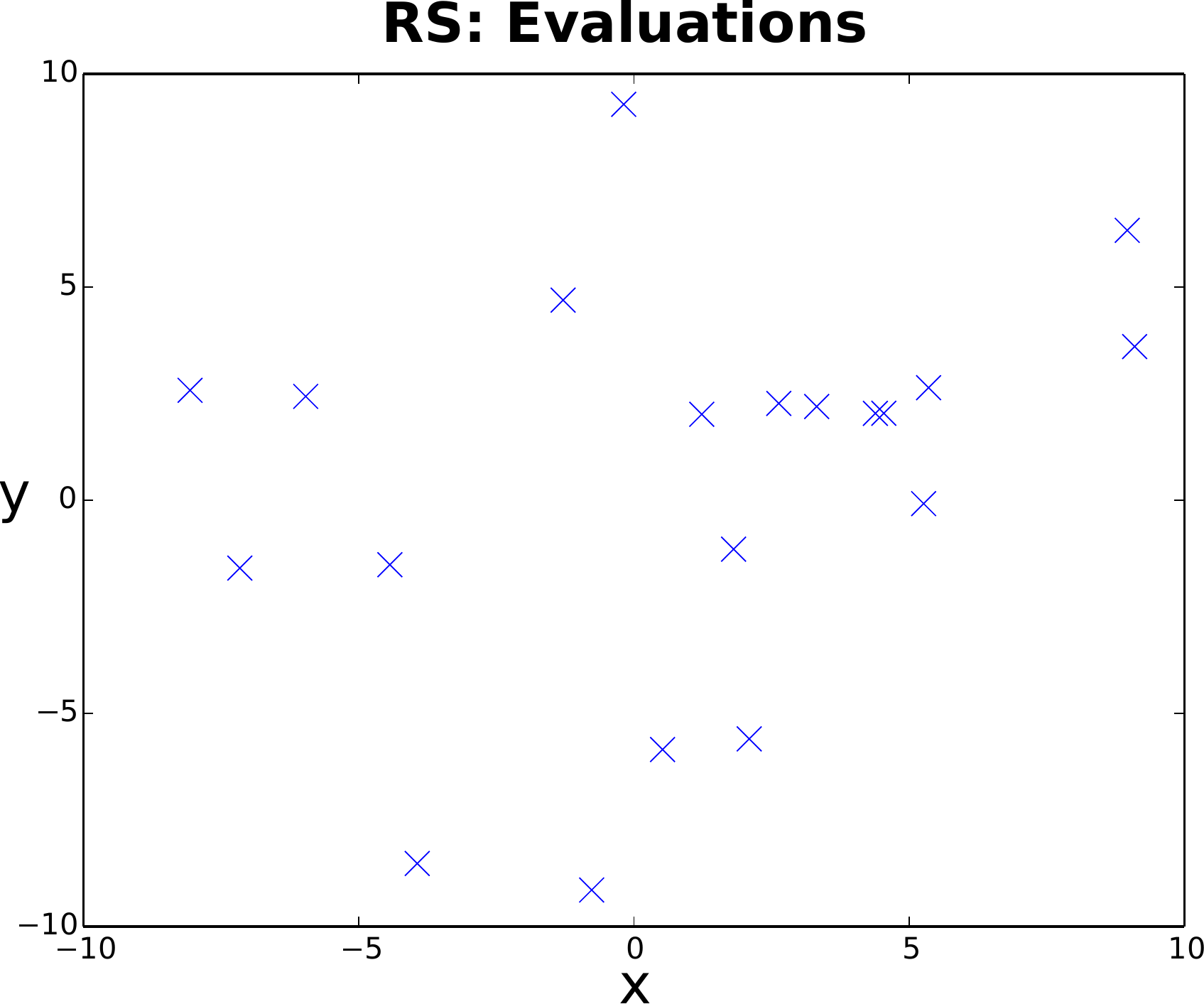}
	\end{tabular}
	\caption{Location in input space (denoted with a blue cross) of each of the evaluations made by PESMOC 
		(top) and RS (bottom). In the case of PESMOC, we also plot the level curves of the 
		acquisition function. Best seen in color. Recall that the feasible region is the box $[0,10] \times [0,10]$.}
	\label{fig:evaluations}
	\vspace{-.5cm}
\end{figure}

To illustrate the usefulness of the acquisition function of PESMOC, we consider the following toy 
2-dimensional optimization problem in the box $[-10,10]\times[-10,10]$:
\begin{align}
\underset{\mathbf{x} \in \mathcal{X}}{\text{min}} & \quad f_1(x,y) = xy, \quad f_2(x,y) = -yx \nonumber\\
\text{s.t.} & \quad x \geq 0, y \geq 0\,.
\end{align}
We note that the feasible space $\mathcal{F}$ is given by the box $[0,10]\times[0,10]$. 
We evaluate PESMOC and RS in this problem and record the evaluations performed by each method at each iteration.
Fig. \ref{fig:evaluations} shows the location  of the evaluations done by PESMOC (top) and RS (bottom) 
in input space, after 20 evaluations of the objectives and the constraints. We observe that PESMOC quickly identifies 
the feasible space, \emph{i.e.}, the box $[0,10] \times [0,10]$, and focuses on evaluating the objectives 
and the constraints in that region. By contrast, RS explores the space more uniformly and, in 
consequence, evaluates the objectives and the constraints more frequently in regions that are 
infeasible. Fig. \ref{fig:evaluations} (top) also shows the level curves of the acquisition function computed
by PESMOC. This function takes high values in regions inside $\mathcal{F}$ and low values 
in regions outside $\mathcal{F}$. These results explain why PESMOC is able to outperform RS in the previous experiments.

\section{Conclusions and Future Work}

We have described an information-based approach that can be used to address a wide range of Bayesian optimization
problems, including multiple objectives and several constraints. Motivated by the lack of methods that are available solve these 
problems with an adequate exploration-exploitation balance, PESMOC has been presented. At each iteration, PESMOC evaluates 
the objective functions and the constraints at an input location that is expected to reduce the entropy 
of the posterior distribution of the Pareto set the most. We have illustrated with synthetic experiments the benefits
of such an approach with respect to a random search strategy. The results obtained show that PESMOC is able to provide 
estimates of the Pareto set of the optimization problem that are more accurate with a smaller number of evaluations.
This is very useful in practical situations in which the objectives and the constraints are very expensive to evaluate.

Future work will compare the proposed method, PESMOC,  with the method described by Feliot \emph{et al.}
\cite{feliot2015bayesian}. Such a method can, in principle, also consider multiple objectives and several constraints. However,
it is based on the expected improvement of the hyper-volume, and in the un-constrained setting there is empirical evidence
supporting that information-based approaches perform better \cite{hernandez2016}. Furthermore, the computation of the
expected improvement of the hyper-volume is very expensive and it can only be done in the case of a small number of objectives,
two or three at most. When the number of objectives is larger, one has to resort to approximate methods that are 
expected to lead to even worse results. 

We also plan to carry out more exhaustive experiments and to include real-world Bayesian optimization problems to assess 
the performance of PESMOC and compare it with the performance of the method described in \cite{feliot2015bayesian}. 
The potential benefits of a decoupled evaluation of the objectives and the constraints will also be explored \cite{hernandez2016}. 
In particular, one should expect increased benefits of such an evaluation setting because the number of functions to 
evaluate (objectives and constraints) is very big in the constrained multi-objective case.

Extra future work may also consider extending the proposed approach, PESMOC, to the setting considered by Shah and Ghahramani 
in \cite{shah2015parallel}. In that setting one does not have to choose a single next location $\mathbf{x}$ on which to
evaluate the objective, but a set of points. Such a batch selection approach may be useful when we have access to a 
distributed system that can evaluate in parallel the objectives and the constraints at several candidate locations.

\section*{Acknowledgments}

The authors acknowledge the use of the facilities of Centro de Computaci\'on Cient\'ifica (CCC) at Universidad Aut\'onoma de Madrid, and  
financial support from the Spanish Plan Nacional I+D+i, Grants TIN2013-42351-P and TIN2015-70308-REDT, 
and from Comunidad de Madrid, Grant S2013/ICE-2845 CASI-CAM-CM.

\label{Bibliography}

\bibliographystyle{acm}
\bibliography{references}

\end{document}